\documentclass[11pt]{article}

\usepackage[final]{acl}

\usepackage{times}
\usepackage{latexsym}
\usepackage{amsmath}
\usepackage{xcolor}

\usepackage[T1]{fontenc}

\usepackage[utf8]{inputenc}

\usepackage{microtype}

\usepackage{inconsolata}
\usepackage{booktabs}
\usepackage{amsmath}
\usepackage{amssymb}

\usepackage{graphicx}
\usepackage{subcaption}

\title{A Circuit for Plural Reference: How LLMs Represent and Retrieve Singular and Plural Entities}

\author{Anh Dang\textsuperscript{1}, 
        Rick Nouwen\textsuperscript{1}, 
        Massimo Poesio\textsuperscript{1,2} \\
        \textsuperscript{1}Utrecht University,
        \textsuperscript{2}Queen Mary University of London
        \\
        \href{mailto:t.t.a.dang@uu.nl}{t.t.a.dang@uu.nl},
        \href{mailto:r.w.f.nouwen@uu.nl}{r.w.f.nouwen@uu.nl},
        \href{mailto:m.poesio@uu.nl}{m.poesio@uu.nl}}

\begin{document}
\maketitle
\begin{abstract}
Coreference resolution is an important task in contextual reasoning. In this paper, we investigate the mechanism for representing and retrieving singular and plural entities for plural reference. We use a combination of mechanistic interpretability and attention pattern analysis to study the process in which LLMs predict a pronoun to refer back to previously mentioned entities. Using a range of causal intervention techniques, we find a set of attention heads that are responsible for (1) representing coreference information in the input, (2) identifying entities that form a plural reference, (3) transferring the information to the component that is responsible for selecting the antecedents and predicting the pronoun. We also find that LLMs align with humans in preference for plural pronoun. Specifically, entities in a plural construction are more likely to be referred to as a plural entity if they are ontologically similar and are linked by the conjunction \textit{and}. \footnote{Our code and data will be made available at \url{https://github.com/dangthithaoanh/plural-reference-circuit}}

\end{abstract}

\section{Introduction}

Discovering the underlying mechanism behind 
Large Language Models (LLMs) has been a topic of 
much research 
in the past few years due to  high interest in improving their safety and their alignment with human behaviors \cite{bereska2024mechanistic, ferrando2024primer}. 
While 
LLMs are becoming larger and more capable, much of their inner working remains under the hood. Understanding their working 
is 
extremely important in making these systems more controllable and thus increasing their safety and explainability.

\begin{figure}[t]
    \centering
    \includegraphics[width=\columnwidth]{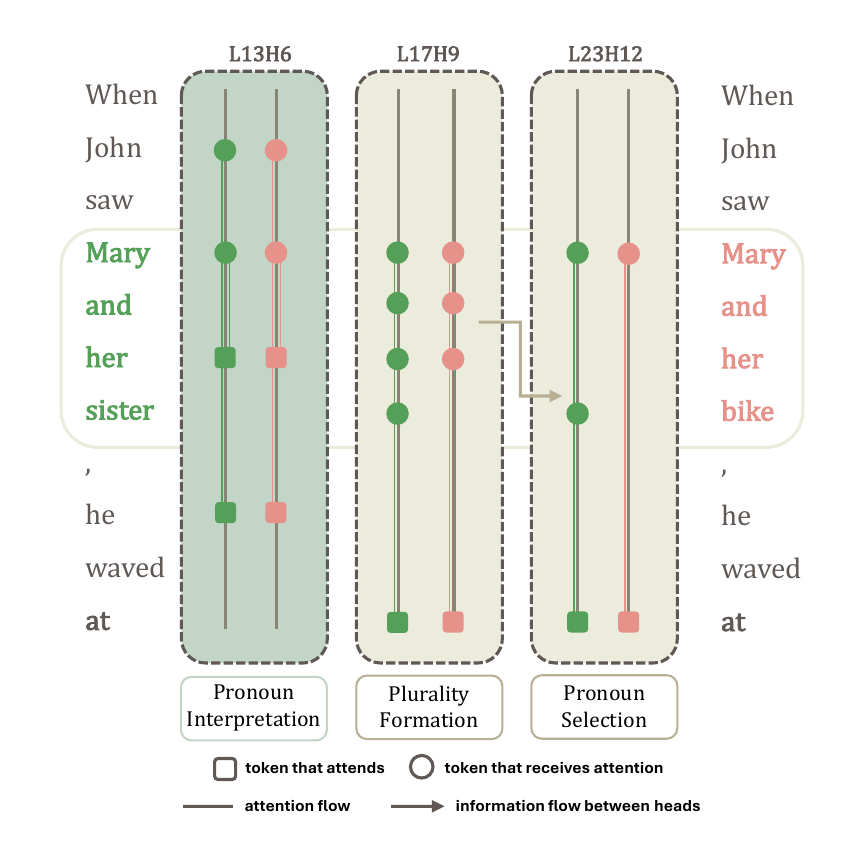}
    \caption{\textbf{Summary of our Plural Reference Circuit}. We identify \textit{Pronoun Interpretation Head}, where entities attend to previous mentions (\textit{he} $\rightarrow$ \textit{John}; \textit{her} $\rightarrow$ \textit{Mary}). \textit{Plurality Formation Head} finds the plural construction that forms a plural entity and attends to them. In contexts where a plural entity is not preferred (\textit{Mary and her bike}), attention to the plural construction is incomplete. Such representation is then sent to the \textit{Pronoun Selection Head}, where the LLMs decide which pronoun to predict and attend to the corresponding antecedents.}
    \label{fig:layer}
\end{figure}

In this paper, we focus on studying how LLMs represent plural reference and how they resolve anaphoric reference involving plurals, as in \textit{John and Mary came to the shop. They bought some milk}, where \textit{they} is understood to corefer with \textit{John and Mary}. In order to resolve plural reference, the LLMs have to identify the set of entities that create a plural entity. 
In psycholinguistic research, processing plural reference has been shown to be more complicated than singular reference because it requires tracking multiple entities simultaneously and grouping the entities that should be co-referred to by a plural pronoun \cite{patson2014processing}. It has been found that entities which are similar in ontological  properties \citep{koh2002resolution} or 
play a similar role 
\citep{moxey2012representing} are more likely to be referred to 
with plurals 
than those which are not. 
On the other hand, entities of the same kind and  linked by a conjunction  are sometimes referred to using a singular pronoun when they create another object, as in \textit{Hook up the engine and the boxcar, and send it to Avon} \citep{cokal2023anaphoric}.

To study 
plural pronoun resolution, 
we use a combination of interpretability techniques, including activation patching, path patching \citep{vig2020investigating, geva2023dissecting, wang2022interpretability}, and attention pattern analysis \citep{wang2022interpretability,tigges2023linear}. We present the flow of information about plural reference across different model components (i.e., embeddings, attention heads) and throughout layers. We first use mechanistic interpretability techniques to find the attention heads that contribute to the LLMs' choice of singular and plural pronouns. We then examine the attention pattern of these components to see what information is used to make the decision. We found a relatively small and 
dense 
circuit, consisting of three groups of heads that are responsible for (i) detecting and encoding plurality signal, (ii) routing this information to the last token position, (iii) identifying the antecedents and producing a plural pronoun to refer to them. A summary of the responsible heads and their functions is provided below. 

\begin{itemize}
    \item \textbf{Pronoun Selection Heads} function at the last token position. They attend to all of the discourse entities to be referenced by the plural/singular pronoun and directly affect the model's decision.

    \item \textbf{Plurality Formation Heads} identify the entities that form a plural entity and send information to the Pronoun Selection heads.

    \item \textbf{Pronoun Interpretation Heads} act as a representation for coreference information where tokens referring to the same entity attend to each other. 
    
\end{itemize}

We also derive the following observations about how LLMs encode referential information and use it to refer back to previous entities: (1) LLMs' attention concentrates on the entities that they refer to; (2) In accordance with previous psycholinguistic experiments, there are constraints on whether a set of entities is a good candidate for plural reference for LLMs. Firstly, LLMs prefer entities with similar ontological properties. For example, the conjoined noun phrase \textit{John and Mary} is more likely to be represented as a plural entity than \textit{John and his bike}. Secondly, preference for plurality is also affected by the linking conjunction. A plural entity is available if the conjunction is \textit{and} (\textit{John and Mary}) and less preferred when \textit{with} is used (\textit{John with Mary}).

\section{Background and Related Work}

In this section, we first survey literature on \textit{coreference resolution} and, more specifically, \textit{plural reference}. We then introduce \textit{mechanistic interpretability}, on which our methodology is based. Finally, we provide a brief explanation about the \textit{Multi-Head Self-Attention} mechanism, the component that we investigate throughout the paper.

\subsection{Coreference Resolution} Identifying previously mentioned entities and referring to them is crucial in language interpretation \cite{poesio-et-al:ARL23}. 
There is an extensive 
body of research on building coreference resolution algorithms \cite{lee(kenton)-et-al:EMNLP17,yu2020free, bohnet2023coreference, martinelli2024maverick}, benchmarking coreference systems \cite{pradhan2011conll,luo&pradhan:anaphora-book:evaluation,recasens&pradhan:ana-book-campaigns}, and evaluating the capabilities of LLMs in resolving coreference \cite{le2023large, upadhye2020predicting, schuster2022sentence, lam2023large,gan-et-al:LREC24:LLM-coref}. 
However, most work on evaluating coreference resolution systems is behavioral and mainly focuses on singular reference \cite{chen2016character, pradhan2011conll}. Much less 
is known about 
how 
modern language models track 
plural 
entities 
throughout the discourse and retrieve them \cite{yu2020free, paun2023scoring}.


In this paper, we track LLMs' decision making process when they have to decide whether to predict a singular or plural pronoun using datasets built on previous psycholinguistic findings on plural reference processing. In the next section, we introduce the structure of plural entities and survey studies on how humans formulate and process plural reference.


\subsection{Plural Reference}

Plural reference is when a plural noun phrase refers to one or more entities. 
In the example mentioned above (i.e., \textit{John and Mary came to the shop. They bought some milk.}), the pronoun \textit{they} is used because the  antecedents \textit{John} and \textit{Mary} are in a plural construction, created by linking the two mentions using the conjunction \textit{and}. This is one way in which a plural entity can be established in discourse. 
However, the presence of the conjunction \textit{and} does not always make a plural entity salient, and its absence does not always prevent the plural entity from being created. For example, it is possible to rephrase the above sentence as \textit{John came to the store with Mary. They bought some milk} while still allowing plural reference \cite{clifton2016discourse}. 
(Cases where antecedents of a plural pronoun are not mentioned together within a conjunction are called \textit{split-antecedent} plural reference.) 

On the other hand, there are also several preferences 
as to whether the entities within a plural construction can be considered a plural entity. 
One of those is ontological similarity \cite{koh2002resolution}. If the entities in the plural construction belong to the same ontological category, they are more likely to be referred to as a plural entity, as found in psycholinguistic experiments. For example, consider the sentence \textit{Mary and her dog came to the shop}. While the conjunction \textit{and} is present and the two mentions are perfectly within a plural construction, it is cognitively easier to continue the sentence with \textit{She} rather than \textit{They}. The formation of a plural entity is more difficult when the two mentions are not ontologically similar. In this paper, we focus only on plural constructions that are conjoined noun phrases (e.g., \textit{John and Mary}), where the two antecedents are connected by a conjunction and are in the same noun phrase. 

In NLP, not much research has been done on how LLMs process plural reference. With respect to coreference ambiguity, \citet{anh2025can} use behavioral methods to study whether LLMs can recognize that a pronoun is ambiguous, meaning that there are multiple candidate referents. They found that LLMs' responses are very sensitive to prompting approaches when they have to specify a referent for a pronoun or decide whether the pronoun is ambiguous, which suggests potential inconsistency in their internal knowledge and the ability to verbalize it. \citet{liu2023we} also find that LLMs are incapable of capturing the ambiguity of the pronoun. Given the gap in our understanding about whether LLMs actually recognize ambiguity and represent it, a natural next question would be: \textit{How do LLMs represent entities that they want to refer to, especially when a plural construction is introduced?} We thus study the inner states of LLMs when they have to predict a pronoun to refer to entities in the plural construction using mechanistic interpretability methods, introduced in the next section.
 
\subsection{Mechanistic Interpretability}
Mechanistic interpretability (MI) is a very active line of work aiming to rigorously explain the inner workings of LLMs \cite{olah2020zoom, elhage2021mathematical}. 
In high-level terms, it seeks to \textit{reverse engineer} the computations of language models into human interpretable processes by attributing the prediction of the model to a sparse set of components at different levels of granularity (e.g., layers, attention heads, neurons). 
Activation patching or causal mediation analysis \citep{vig2020investigating, wang2022interpretability, geva2023dissecting, goldowsky2023localizing, meng2022locating, geiger2021causal, mueller2024quest} is one of the foundational techniques in MI. 
It is a method that allows for estimating the effect of a specific component on models' prediction. 
Given an input sequence, by exchanging the activation of certain tokens that are critical for prediction with a corrupted activation and measuring how much the prediction has changed, activation patching can locate the components that directly or indirectly affect the output. This technique has been used to investigate a range of processes,  from low-level behaviors such as factual knowledge retrieval \cite{meng2022locating, geva2023dissecting}, indirect object identification \cite{wang2022interpretability}, entity tracking \cite{dai2024representational, dai2026cell, prakash2024fine}, truthfulness representations \cite{marks2023geometry}, and the \textit{greater-than} computation \cite{hanna2023does} to more complex reasoning processes such as question-answering \cite{basu2025mechanistic, wiegreffe2025answer}. MI techniques allow  intervention on a wide range of model components. In this paper, we study the activity of attention heads, the component that is found to be responsible for capturing entity relations in previous studies. We briefly explain the attention mechanism in the next section.




\subsection{Self-Attention Mechanism}

We provide background on the architectural detail of the self-attention mechanism, the component that we exclusively focus on in this paper. We investigate auto-regressive Transformer-based LLMs \cite{vaswani2017attention}. Given a sequence of input $X$ consisting of $n$ tokens $T = (t_0, t_{1}, ..., t_{n-1})$, the model builds the representations $x_i$ of each token $t_i$ at each layer $l$: $X^l = (\mathbf{x}^l_0, \mathbf{x}^l_{1}, ..., \mathbf{x}^l_{n-1}  $). At layer $l$, the representation $X^l$ is processed by a Multi-head self-attention module (MHSA) and an MLP module. The output resulting from both operations is written in the \textit{residual stream} \cite{elhage2021mathematical}, which serves as the channel of information flow across the model components.   

The MHSA modules weigh the importance of each token to each other. Each head assigns a set of attention weights for each pair of tokens. The output of a head is computed using the \textit{Key} ($K$), \textit{Query} ($Q$) and \textit{Value} ($V$) vectors \cite{elhage2021mathematical}. For a pair of tokens ($t_i, t_j$), the attention weight from the current token $t_j$ to a previous token $t_i$ is computed using the $QK$ matrix.  The $OV$ matrix at token $t_j$ decides what information of $\mathbf{x}_j$ is written to the residual stream. The $QK$ and $OV$ circuits operate independently from each other. 

Recent LLMs such as the \texttt{Qwen3} \cite{yang2025qwen3}  and \texttt{Llama3.2} \cite{grattafiori2024llama} model families are trained using \textit{grouped-query attention} \cite{ainslie2023gqa}. In the original MHSA architecture, every query head has distinct key and value matrices. GQA reduces the computation cost by having  multiple query heads share the same key-value head. Since we focus on these model families in our paper, we use GQA in our explanation of the MHSA module. 

Let $h$ denote the attention ($Q$) head and $k$ denote the key-value ($KV$) head. Under GQA, each query head $h$ is assigned to a KV head. The projection matrices are 
defined as follows: $\mathbf{W}^h_Q \in \mathbb{R}^{d_{\text{model}} 
\times d_{\text{head}}}$ and $\mathbf{W}^h_O \in \mathbb{R}^{d_{\text{head}} 
\times d_{\text{model}}}$ are unique to each query head $h$, while 
$\mathbf{W}^k_K \in \mathbb{R}^{d_{\text{model}} \times d_{\text{head}}}$ 
and $\mathbf{W}^k_V \in \mathbb{R}^{d_{\text{model}} \times d_{\text{head}}}$ 
are shared across all query heads assigned to KV head $k$. The Key ($\mathbf{k}^k_i$), Query ($\mathbf{q}^h_j$), and Value ($\mathbf{v}^k_i$) vectors are calculated as follows.
\begin{center}
    $ \mathbf{v}^k_i$ = ${\mathbf{W}^k_V}{\mathbf{x}_i}$, $\mathbf{k}^k_i$ = ${\mathbf{W}^k_K}{\mathbf{x}_i}$, $\mathbf{q}^h_j$ = ${\mathbf{W}^h_Q}{\mathbf{x}_j}$
\end{center}
The model calculates the attention weight from the current token $t_j$ to a previous token $t_i$, denoted by $\alpha_{t_j,t_i}$, by passing the dot product of the query ($Q$) at $t_j$ and key ($K$) vector at $t_i$ through a softmax.

\[
\alpha_{ji
} =
\operatorname{softmax}
\left(
\frac{\mathbf{x}_j^\top \mathbf{W}_Q^\top \mathbf{W}_K \mathbf{x}_i}
{\sqrt{d_{\text{head}}}}
\right)
\]

The intermediate output of the query head $h$ at $t_j$, denoted by $\mathbf{z}^{h}_j$, is the linear combination of value vectors and the attention pattern across all attended positions:

\[
\mathbf{z}^{h}_j = \sum_{i=0}^{j} \alpha^h_{ji} \mathbf{v}^k_i
\]

Through the $OV$ circuit, $\mathbf{z}^{h}_j$ is projected and written to the residual stream.

\[
\mathbf{o}^h_j = \mathbf{z}^h_j \mathbf{W}^h_O
\]

The output of the attention module $\mathbf{a}^l_j$ is the sum of $\mathbf{o}^h_j$ across all attention heads.

\section{Dataset and Evaluation}

\subsection{Task Description}

We design a pronoun prediction task to trigger LLMs' mechanism for plural reference resolution. Given the 
sentence prefix \textit{When John saw Mary and her sister, he waved at \_\_}. The model may produce \textit{her} to refer to \textit{Mary} or \textit{Mary's sister}. 
Or 
the pronoun \textit{them} can be used to refer to the group of \textit{Mary and her sister}.

\subsection{Dataset}

We generate two synthetic datasets of sentence prefixes for our experiment. The first dataset ($D_{pl}$) includes sentence prefixes that introduce plural entities and expect the LLMs to predict a plural pronoun [1]. The second dataset ($D_{sg}$) also introduces a plural construction, but a plural pronoun is dispreferred [2]: 

\textcolor[HTML]{E2A16F}{[1] When John saw \textbf{Mary and her sister}, he waved at \_\_}. 

\textcolor[HTML]{86B0BD}{[2] When John saw \textbf{Mary and her bike}, he waved at \_\_.}

For each sentence prefix $s_{pl}$ $\in $ $D_{pl}$, there are several discourse entities, namely the subject \textit{John} ($s$), \textit{Mary} ($e_1$), and \textit{her sister} ($e_2$). The plural construction consists of three elements, which are \textit{Mary} ($e_1$) and \textit{her sister} ($e_2$) and is formed by the conjunction \textit{and} ($c$). The plural construction can be formalized as a tuple of $(e_1, c, e_2)$. Since $e_1$ and $e_2$ form a plural entity, the plural pronoun \textit{them} ($r_{pl}$) is more likely to be used. However, it should be noted that in this context, a singular pronoun (e.g., \textit{her}) is also a plausible continuation. 

In $D_{sg}$, we expect LLMs to predict a singular pronoun while retaining the original structure. We replace the second element of the plural construction ($e_2$) with an inanimate entity $e^{'}_2$ (\textit{her bike}) while keeping the rest of the sentence prefix unchanged. We carefully select a set of verbs that are not compatible with inanimate entities (e.g., \textit{laughed at}, \textit{nodded at}) so that it is semantically infelicitous to produce a plural pronoun. As such, the model is expected to generate a singular reference ($r_{sg}$), referring only to \textit{Mary} ($e_1$). 

We generated 300 prefixes for each dataset. All entities are single-token and all prefixes are equal in length. The gender of the entities is always randomized. For example, the subject $s$ can be either male or female and $e_1$ and $e_2$ may have the same or different gender. See Table ~\ref{data} for information on our templates.


\subsection{Evaluation} 
Before finding the mechanism for plural reference resolution, we test how LLMs perform the pronoun prediction task. We are particularly interested in their preference in choosing the pronoun when a plural construction is present, especially with different gender, ontological similarity \cite{clifton2016discourse}, and also linking conjunctions \cite{moxey2012representing}. We use $D_{pl-sg}$ to indicate the probability difference between the plural and singular pronoun, 
and introduce three variables:

\begin{itemize}
    \item \textbf{Gender:} $e_1$ and $e_2$ can have the same or different gender: (Mary, and, Emma), (Mary, and, John)
    \item \textbf{Ontological Similarity}: $e_1$ and $e_2$ antecedents have the same or different ontological properties. We consider here three levels of similarity, based on noun type (proper or common) and animacy (animate or inanimate): (Mary, and, Emma) > (Mary, and, her sister) > (Mary, and, her bike)
    \item \textbf{Conjunction}: The conjunction linking $e_1$ and $e_2$ can be \textit{and} (Mary, and, Emma) or \textit{with} (Mary, with, Emma). Previous psycholinguistic studies show that the conjunction \textit{with} \footnote{By definition, \textit{with} is a preposition, not a conjunction. We chose to use the conjunction-like function of \textit{with} for the purpose of constructing the counterfactual pairs for the patching experiments} decreases the preference for choosing a plural pronoun over a singular one \cite{moxey2004constraints, moxey2012representing, albrecht1998accessing, sanford1990description}. 
\end{itemize}

Fitting 
Linear Mixed-Effect Regression models to estimate the effect of these three variables on the probability difference between $r_{sg}$ and $r_{pl}$ ($D_{pl-sg}$),
we replicate 
the effects of ontological similarity and conjunction that are expected from the psycholinguistic literature. 
The less ontologically similar the two antecedents are, the less likely they are to be referred to as a plural entity. In addition, a plural pronoun is less preferred when the antecedents are linked by \textit{with}, compared to \textit{and}. (See Appendix ~\ref{b} for detailed results for all investigated models.)

This analysis 
guided 
our patching experiment,
as 
we chose the type of plural construction that does not always trigger the formation of plural reference. The motivation behind our choice is the similarity in structure and token length between $s_{pl}$ and $s_{sg}$. As such, to ensure the validity of the patching results, we extract the prediction for $D_{pl}$ and only retain the prompts where the model predicts the plural pronoun.  

\begin{figure}[t]
    \centering
    \includegraphics[width=\columnwidth]{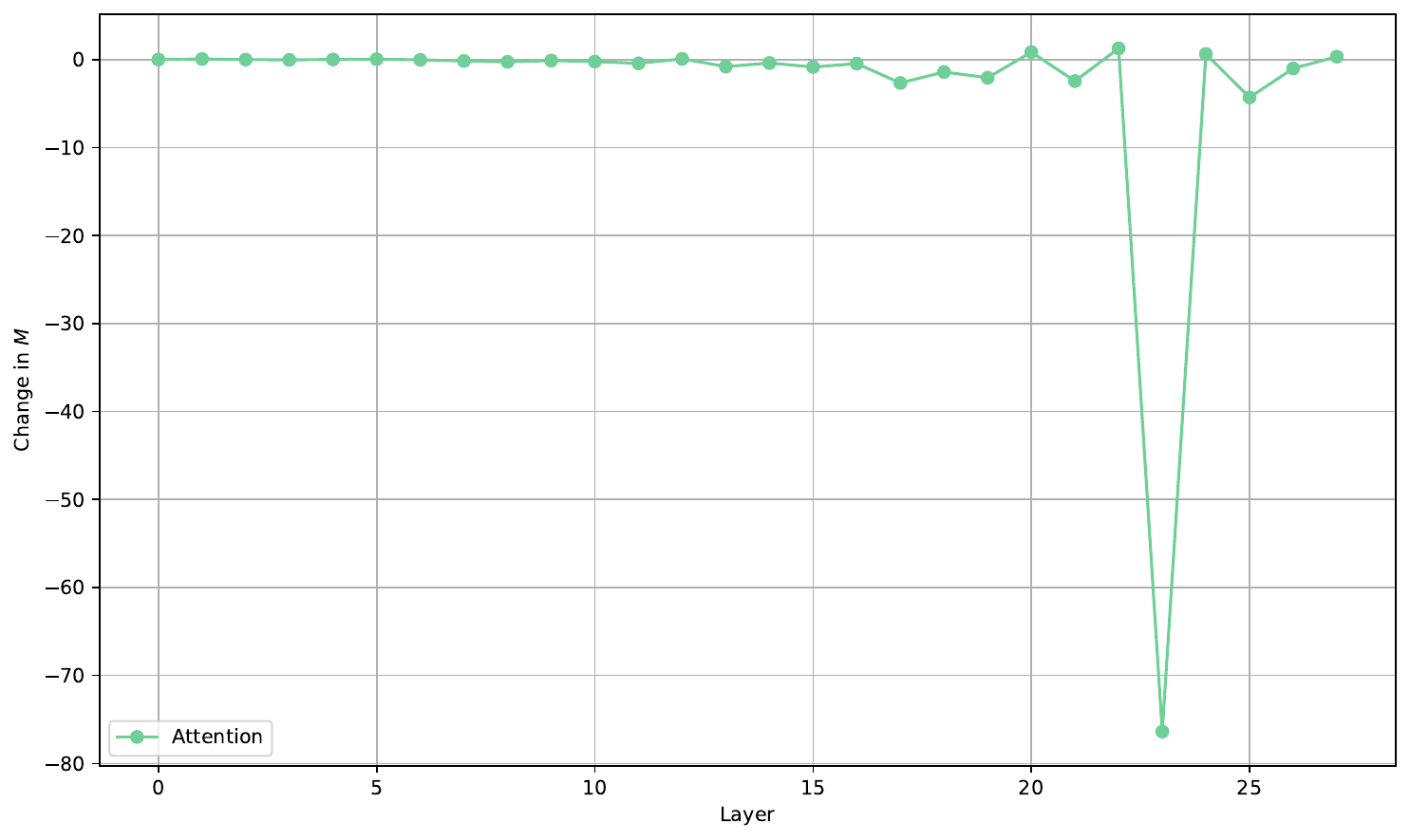}
    \caption{Layer-wise direct effect of MHSA modules on $M$ }
    \label{fig:layer}
\end{figure}

\section{Experimental Setup}

\begin{figure*}[ht]
    \centering
    \includegraphics[width=\linewidth]{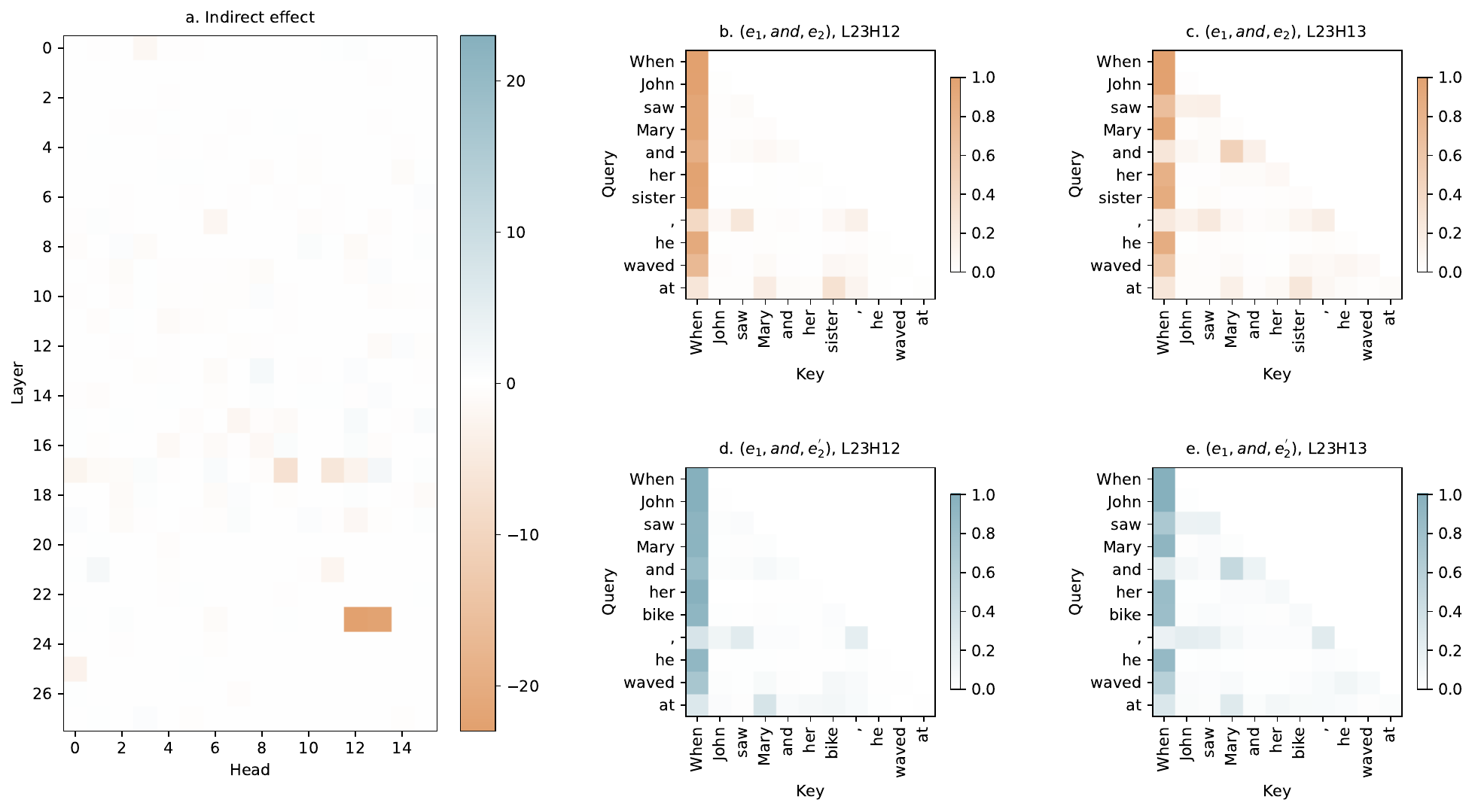}
    \caption{\textbf{Overview of intervention results and attention pattern visualizations of major heads.} (a) Indirect effect of individual head intervention across layers. \textcolor[HTML]{E2A16F}{Brown} values denote negative change in $P_{pl}$. (b) and (c) Attention pattern of Pronoun Selection heads for $D_{pl}$ (\textcolor[HTML]{E2A16F}{brown}). (d) and (e) Attention pattern of Pronoun Selection heads (\texttt{L23H12 and L23H13)} for $D_{sg}$ (\textcolor[HTML]{86B0BD}{blue}).   }
    \label{fig:main1}
\end{figure*}


\paragraph{Experiment}

We use a combination of \textit{activation patching} and \textit{path patching} in our experiment. Activation patching intervenes on a component $C$ while allowing all downstream components to be affected. It measures the \textit{indirect} effect of $C$ on the output. \textit{Path patching} involves another step, restoring the activations of all downstream components to their original values \cite{wang2022interpretability, goldowsky2023localizing, meng2022locating}. The goal is to isolate the direct effect of that component. If path patching causes a decrease in $P_{pl}$, the direct effect can be attributed to that component \cite{pearl2022direct}.

To perform path patching on the component $C$, which can be high-level units such as the residual stream of a specific layer ($\mathbf{x}^l_i$) or lower-level ones such as the value vector of a specific token at a head ($\mathbf{v}^h_i$), we perform the following steps. 

\begin{itemize}
    \item \textbf{Original Run} Pass $s_{pl}$ to the model and record the original probability of the plural pronoun $P_{pl, ori}$. 
    \item \textbf{Corrupted Run} Run a forward pass through $s_{sg}$ then cache the activations of all token positions at each layer ($\mathbf{X}^{l}$).
    \item \textbf{Intervened Run} Replace the activation of $C$ at position \textit{i} with that at the corresponding position of $s_{sg}$, 
    restore the activations of all upstream components to their clean values to block any indirect propagation of the corrupted signal through the network, run the forward pass through $s_{pl}$ again, and recompute $P_{pl, inter}$.
\end{itemize}

By patching, we intentionally replace the attention output of the patched token with that of the corrupted prompt ($e_2$), where the second mention is an entity that is less likely to be referred to together with the first. If there is a set of heads that is responsible for searching for the antecedent to be referenced, intervening on its attention output would lead to a change in prediction. As such, if intervening on a specific head causes $P_{pl}$ to decrease after intervention, it can be causally inferred that it is one of the heads responsible for pronoun resolution. 

\paragraph{Constructing Corrupted Prompts} Our regression models show that ontological similarity and conjunction significantly affect the preference to choose a plural pronoun. We treat them as crucial factors that either enhance or corrupt the plurality signal. As such, we use two corrupted prompts, manipulating either $e_2$ or $c$: ($e_1$, c, $e^{'}_{2}$) and ($e_{1}$, $c'$, $e_2$). 

\paragraph{Intervening Location} For each component, we run the intervention at multiple positions, spanning from the last token up to the position of the corrupted token. We find that only the intervention on the \textit{last token} and \textit{corrupted token} ($e_2$ and $c$) show a considerable effect. As such, all the findings presented in the following sections arise from intervening on these two positions.    

\paragraph{Metrics} We measure the effect of the intervention by calculating how much $P_{pl}$ changes after the intervention. We define the evaluation metric $M$ as the percentage of probability difference between $P_{pl, ori}$ and $P_{pl, inter}$. Negative values of $M$ mean that the intervention suppresses the plurality signal and thus decreases $P_{pl}$.

\[
M =
\left(
\frac{P_{\text{pl, inter}} - P_{\text{pl, ori}}}
{P_{\text{pl, ori}}}
\right)
\times 100
\]

\paragraph{Models} We patch on different sizes of the \texttt{Qwen3} family (\texttt{Qwen3-0.6B, Qwen3-1.7B}) and \texttt{GPT2} (\texttt{GPT2-medium}). We present in the main text the circuit of \texttt{Qwen3-1.7B}. Comparable results of other  models can be found in Appendix ~\ref{e}.

\begin{figure*}[h]
    \centering
    \includegraphics[width=\linewidth]{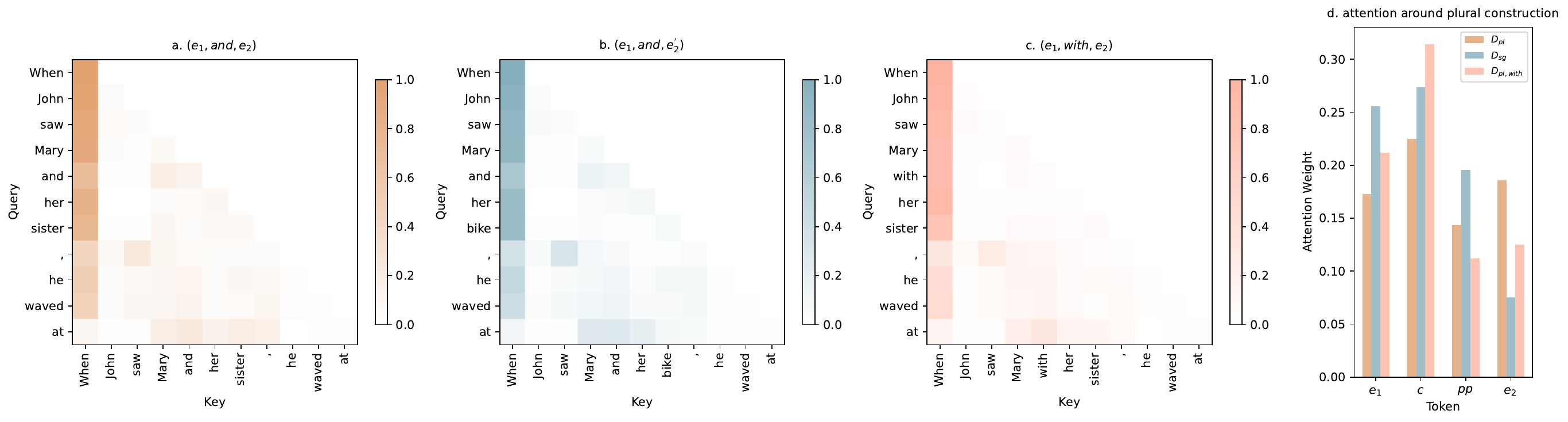}
    \caption{\textbf{Attention Pattern of Plurality Formation Head for $D_{pl,a}$, $D_{sg,a}$, and $D_{pl,w}$  } (a), (b) and (c): Attention pattern of the Plurality Formation head for $D_{pl,a}$, $D_{sg,a}$, and $D_{pl,w}$. (d) Numerical values of attention weights on each token in the plural construction, where $e_1$ and $e_2$ are the two entities, $c$ is the conjunction and $pp$ is the possessive pronoun of $e_2$}
    \label{fig:your2}
\end{figure*}

\section{The Plural Reference Circuit} 

In this section, we present our findings of the coreference circuit, including the core components and their functionalities. We first performed path patching on the MHSA module at the layer level to find the layers that directly affect $P_{pl}$ (Figure ~\ref{fig:layer}). We then proceed to patch the components comprising them, namely the individual attention heads. We run activation patching on each attention head to estimate their indirect effects on the MHSA module at Layer 23. As we patch at each position, we select the set of heads that cause the largest decrease in probability of the plural pronoun (Figure \ref{fig:main1}a).

\paragraph{Pronoun Selection Heads identify candidate antecedents}  Figure ~\ref{fig:layer} shows that the MHSA module at Layer 23 is the only one that directly affects the output. With activation patching, we find that heads 12 and 13 of layer 23 (hereafter \texttt{L23H12} and \texttt{L23H13}) have the strongest effect on $M$ and are both responsible for the prediction process of the pronoun. As the first step toward understanding the function of these heads, we investigate their attention weights. Figure \ref{fig:main1}b (top, middle) shows the mean attention weight of \texttt{L23H12} across all prefixes in $D_{pl}$. It can be seen that the last token (\textit{at}) attends the most to the first ($e_1$) and second entity ($e_2$) in the plural condition, which implies that they are considered to be the antecedents of $r_{pl}$. Notably, we observe that $e_2$ receives more attention than $e
_1$. In $D_{sg}$ ( Figure \ref{fig:main1}d), we observe that the attention weight is given to $e_1$, which is the antecedent of the singular pronoun.

On the other hand, at \texttt{L23H13} (Figure ~\ref{fig:main1}c), while the attention to both antecedents is still present, we also see a very high attention weight between the conjunction \textit{and} and $e_1$. These attention patterns at these heads suggest a strong focus on the elements that are crucial for plural reference resolution, namely antecedents and conjunctions.

These attention patterns suggest 
that 
\texttt{L23H12} and \texttt{L23H13} 
might be called the \textbf{Pronoun Selection} heads: they identify the antecedents to be referred to by the pronoun. 
In a correlation experiment (Appendix ~\ref{c}), we found that the amount of attention difference between $e_1$ and $e_2$ positively correlates with the probability difference between $P_{sg}$ and $P_{pl}$, such that the larger the attention on $e_2$ compared to $e_1$, the larger the difference between $P_{pl}$ and $P_{sg}$. 
This type of head is 
similar 
to the Name Mover Head found in \citet{wang2022interpretability} and other MI studies \cite{prakash2024fine, kim2025reasoning, lieberum2023does}, which identifies and encodes the decisive information for prediction.

\paragraph{Plurality Formation Heads send information about the plural entity to Pronoun Selection Heads}

We just saw how 
the Pronoun Selection Heads
affect the prediction
by attending 
to the discourse entities that are the candidate antecedents of the predicted pronoun. 
We now ask what makes the model attend more to $e_2$ in $s_{pl}$. 
To answer this question, 
we decompose the effect of the $K$, $Q$, and $V$ vectors of the Pronoun Selection head. Since the attention weight is computed using the dot product of the vectors $K$ and $Q$, we first identify which is more crucial for prediction. We run separate interventions on these vectors and measure $M$. For the $Q$ vector, we patch at the last token position, because here it decides which token it needs to attend to. For the $K$ and $V$ vectors, we patch at $e_2$, where information is offered to the query vector. We found that the $Q$ vector does not affect the prediction, while patching the $K$ and $V$ vectors causes a strong drop in $P_{pl}$.

\begin{figure}[t]
    \centering
    \includegraphics[width=\columnwidth]{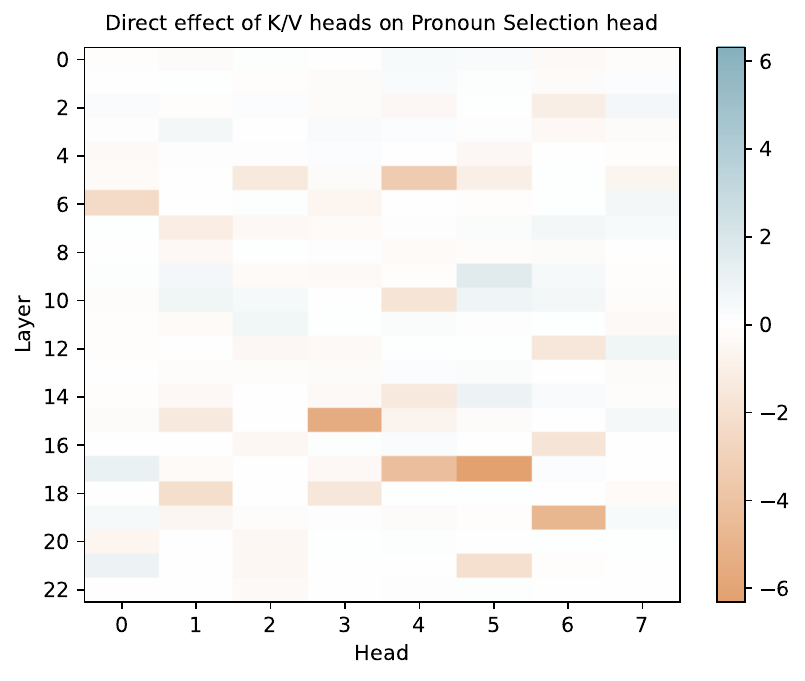}
    \caption{Direct effect of K/V heads on Pronoun Selection head, measured by change in $M$}
    \label{fig:your-label}
\end{figure}

Next, we ask which head affects the Value vector of the Pronoun Selection head. We intervene on the path between the $V$ vector of the Pronoun Selection head and other $K/V$ heads. The value vector at $e_2$ ($\mathbf{v}^k_{e_2}$) is replaced with $\mathbf{v'}^k_{e'_2}$ from $D_{sg}$. The output $\mathbf{o'^h_j}$ of head $h$ after the intervention is updated as follows.    
\[
\mathbf{z'}^{h}_{j} = \sum_{i=0}^{j} \alpha^h_{j,e_2} \mathbf{v'}^k_{e'_2} ;
\qquad
\mathbf{o'}^h_j = \mathbf{z'}^h_j \mathbf{W}^h_O
\]

Figure ~\ref{fig:your-label} shows that some $K/V$ heads at layers 15, 17, 19, and 21 directly affect the Value vector of $e_2$ at the Pronoun Selection head, which suggests that it reads information from these components. Figure ~\ref{fig:main1} shows that these heads also indirectly affect $P_{pl}$. We inspect the attention pattern of the attention heads that are associated with these $K/V$ heads to find what information they offer to the $V$ vector of $e_2$. Among these heads, \texttt{L17H9} has the strongest effect on the Pronoun Selection head. Figure \ref{fig:your2} shows the attention pattern of \texttt{L17H9} for $D_{pl,a}$,  $D_{sg,a}$ and $D_{pl,w}$. 
For the sake of comparison, we incorporate the type of conjunction into the datasets, $a$ stands for \textit{and} and $w$ stands for \textit{with}. In the case of $D_{pl,a}$, the last token \textit{at} attends to the whole plural construction, with slightly more attention on $c$ and $e_{2}$ (Figure ~\ref{fig:your2}a). This behavior is quite different from the attention pattern of the Pronoun Selection head, where the attention weights are only dominant at $e_1$ and the second token of $e_2$. For $D_{sg,a}$ (Figure ~\ref{fig:your2}b), it only assigns attention weights up to the possessive pronoun of $e_2$ (\textit{her}). Interestingly, when the conjunction \textit{with} is present, the attention is only on  $e_1$ and $c$. The attention weight amount on $e_2$ is almost negligible (Figure ~\ref{fig:your2}c). This result is compatible with our earlier statistical analysis of the conjunction effect, which shows that the LLMs are much more likely to establish a plural entity from a plural construction when the conjunction is \textit{and} rather than \textit{with}. 
Taken together, these patterns suggest that this head is trying to find the set of entities to be referenced by selectively attending to the tokens that fit the constraints for a plural entity. It attends to the whole plural construction when it is a qualified plural entity. We hypothesize that it is responsible for constructing, representing, and writing the information about the plural entity to the Pronoun Selection heads.

\begin{figure}[t]
    \centering
    \includegraphics[width=\columnwidth]{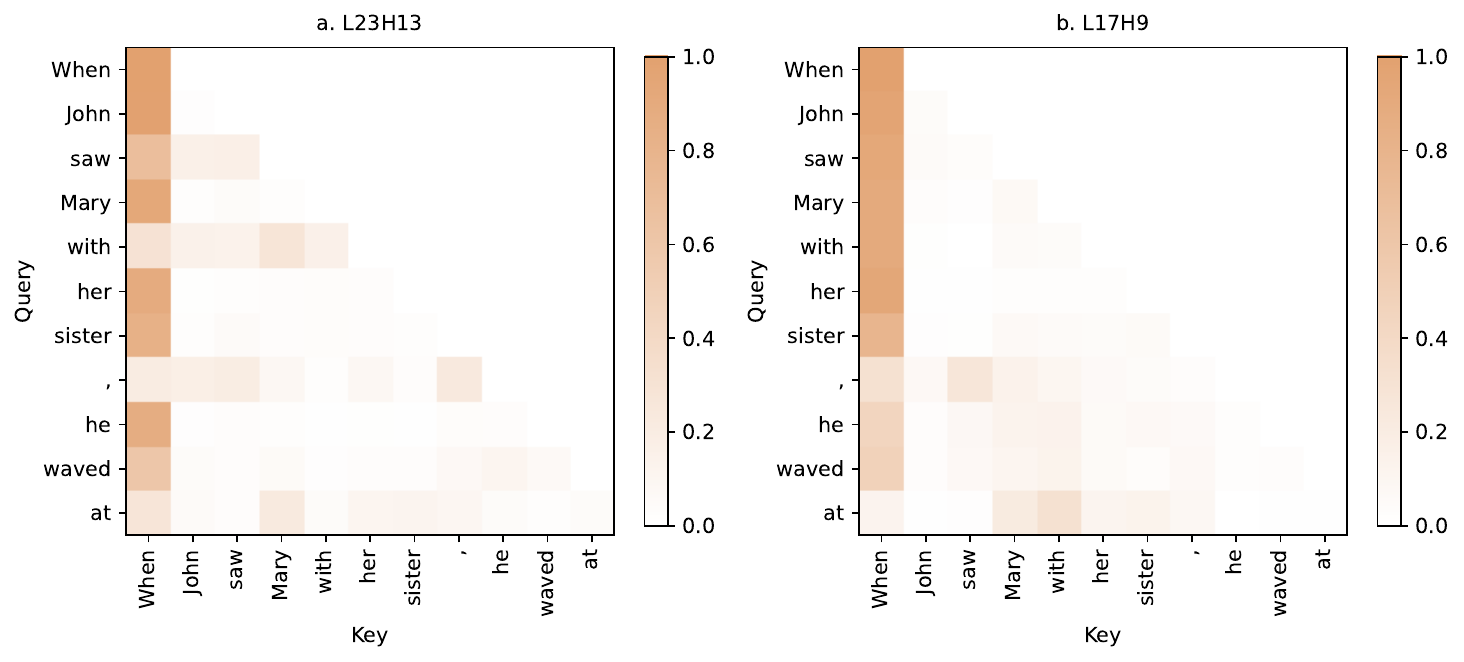}
    \caption{\textbf{Comparing attention pattern of the Pronoun Selection head and the Plurality Formation head}. (a) attention pattern of \texttt{L23H13} for $s_{pl,w}$, (b) attention pattern of head \texttt{L17H9} for $s_{pl,w}$}
    \label{fig:with}
\end{figure}

In addition, in Figures \ref{fig:your2}a and \ref{fig:your2}b, where the conjunction \textit{and} is used, it attends to $e_1$, similar to what we observe in one of the Pronoun Selection heads. This behavior is not present when \textit{with} is used (Figure \ref{fig:your2}c). In Figure ~\ref{fig:with}, we compare the attention pattern of $D_{pl,w}$ at the Pronoun Selection head (Figure ~\ref{fig:with}a) and the  Plurality Formation head (Figure ~\ref{fig:with}b). We found that while \textit{with} does not attend to $e_1$ at the Plurality Formation head, it still attends to $e_1$ at the Pronoun Selection head. There is evidence that some attention heads encode grammatical dependencies \cite{voita2019analyzing, htut2019attention, clark2019does}. Therefore, we speculate that the attention from \textit{with} to $e_1$ is purely syntactic at the Pronoun Selection head and is not directly related to the process of identifying plural entities.  

\begin{figure}[t]
    \centering
    \includegraphics[width=\columnwidth]{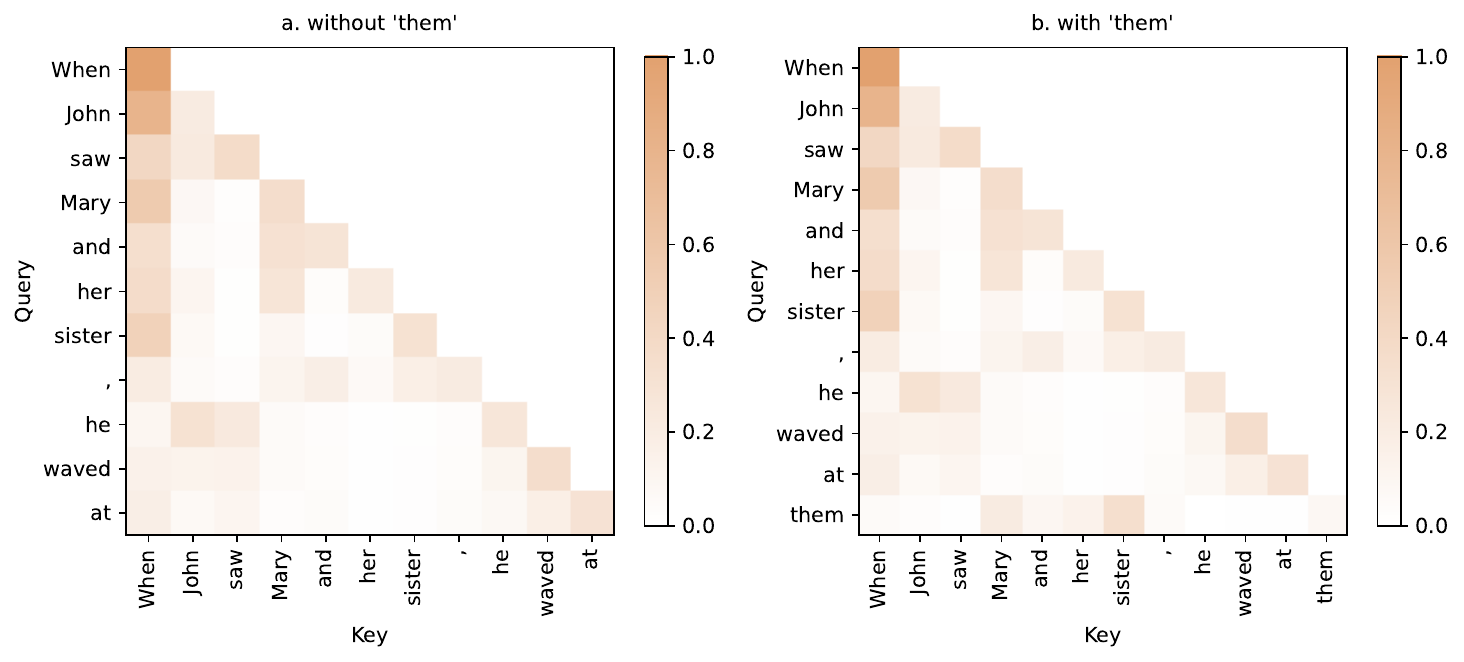}
    \caption{\textbf{Attention Pattern of Pronoun Interpretation head}. (a) attention pattern of \texttt{L13H6} for $s_{pl}$, (b) attention pattern of head \texttt{L13H6} for $s_{pl} + them$ }
    \label{fig:them}
\end{figure}

\paragraph{Pronoun Interpretation Heads encode coreference information of previous tokens} 
Finally, we find a head that encodes coreference information. This head affects around 25\% of the probability difference. When the token is a pronoun, it attends to the antecedent of the pronoun that was mentioned earlier in the input. In Figure ~\ref{fig:them}a, it can be seen that \textit{her} attends to \textit{Mary} and \textit{he} attends to \textit{John}. To verify how this head behaves when there is a plural pronoun, we add \textit{them} to the prompt and inspect the attention pattern (Figure ~\ref{fig:them}b). It can be seen that \textit{them} attends to the whole plural construction (Mary, and, her sister), with much stronger attention on $e_2$ (sister) and the conjunction \textit{and}. 
%
We hypothesize that while this head may not play a direct role in predicting the pronoun, it functions as a coreference map, where every entity attends to its previous mentions. This head was also discovered in earlier studies on the attention pattern of language models \cite{clark2019does, tenney2019bert}.

\paragraph{Generalization to other models} We validate our circuit across other sentence templates and observe the same attention pattern across all heads in the circuit for the variant. We also extract the circuit for other models and find that the circuit generalizes within the \texttt{Qwen3} family and also to \texttt{GPT2-medium} (Appendix ~\ref{e}).

\section{Discussion}
Our patching experiments reveal a mechanism for performing the plural pronoun prediction task. 
In the middle layers, the model first builds up the representation for an entity or a group of entities that it needs to refer back to. 
At the Plurality Formation head, the last token 
before reference production
assigns attention weights spanning the phrase that contains the candidate entities. If the entities align with the model's preference for plural entities (being ontologically related and connected by a preferred conjunction), they will be attended to. 
After deciding on the entity or a set of entities to be referenced, the Plurality Formation heads write this information into the residual stream. The Pronoun Selection head receives this information, attends to the known entities, and uses it for making decisions. In addition, the models dedicate a head for encoding coreference information of all mentioned entities.    

\section{Conclusion}

In this paper, we investigate how LLMs represent and use coreference information for pronoun prediction, especially in the case of plural reference. 
We find the constraints that LLMs use to decide whether to use a singular or a plural pronoun when a plural construction is present. We then use mechanistic interpretability methods, complemented by attention pattern analysis, to extract a circuit that is responsible for 
identifying the entities to be referred to and propagating such information across components to generate the correct pronoun.

\section*{Limitations}

Our study only focuses on the most basic and common type of plural reference (conjoined noun phrases) and plural constructions that have two elements. In actual language use, plural reference is often more complex and the antecedents may not be linked by a conjunction (e.g., split-antecedent reference). Psycholinguistic studies have shown that it is possible and preferred to use a singular pronoun to refer to a plural entity when the antecedents create a new and unified object \cite{cokal2023anaphoric}. It would be interesting to study the model's circuit behavior in these cases. 

In addition, since our goal is to understand the high-level mechanism behind the process of pronoun prediction, we focus more on analyzing the function of each component than extracting a complete circuit for coreference. As such, there is still uncertainty about whether the circuit is fully unique to coreference. We also do not thoroughly investigate the function of MLPs. Although we show that intervening on the MLP module at the late layer strongly affects $P_{pl}$, we know little about its functionality in the circuit.

\section{Ethical Considerations}

The study only uses synthetic datasets generated by LLMs. It does not include any personal data from human participants.

\section{Statement of AI Usage}

During the project, AI was used for coding assistance, dataset generation, and spelling check. Project ideas, experimental setup, interpretation, and writing were done by the authors.

\section{Acknowledgments}

We would like to thank three anonymous reviewers for their insightful comments and suggestions. This project is funded by the Dutch Research Council (NWO) through the AiNed Fellowship Grant (Dealing with Meaning Variation, NGF.1607.22.002) to Massimo Poesio. 


\bibliography{custom}

\newpage

\appendix

\section{Model Configuration}  \label{a}

We find the coreference circuit for a range of model families and sizes. We use two recent and widely-used LLM collections, namely \texttt{Qwen3} and \texttt{GPT2}. These models are the standard choices for most mechanistic interpretability research \cite{wang2022interpretability, meng2022locating}. Table \ref{data2} shows the configurations of the models, including model size, number of layers, number of attention heads, number of key-value heads and the size of hidden activations.

\begin{table*}[t]
\centering
\resizebox{\textwidth}{!}{%
\begin{tabular}{lcccc}
\hline

\textbf{Model} & \texttt{num\_layers} & \texttt{num\_key\_value\_heads} & \texttt{num\_attention\_heads} & \texttt{hidden\_size} \\
\hline
\hline

Qwen3-0.6B & 28 & 8 & 16 & 1024 \\
Qwen3-1.7B & 28 & 8 & 16 & 2048 \\
GPT2-medium & 24 & 16  & 16 &  1024 \\

\hline

\end{tabular}
}
\caption{Configurations for LLMs in the Qwen3 and GPT2 families}
\label{data2}
\end{table*}

\section{Statistical Analysis}   \label{b}

In this section, we report details about our statistical analysis, which studies the effect of gender, ontological similarity, and conjunction preference for choosing plural pronouns. See Table ~\ref{data_t} for details about our dataset. We generate a total of 349 prompts and ensure a balanced distribution between the variables. 

We fit Linear Mixed-Effects Regression models with Restricted Maximum Likelihood (REML) with the verbs as random effects. The models are run using the \texttt{lme4} package in R. Our dependent variable is the probability difference between the plural and singular pronoun.

\vspace{0.2cm}

$prob-diff = P_{plur} - P_{sing} $ 

\vspace{0.2cm}

The independent variables include the following.

\begin{itemize}
    \item \texttt{Gender}: same, different
    \item \texttt{Ontological  (Ont. Sim.)}: N + N (proper name + proper name), N + R (proper name + relation), N + O (proper name + object) 
    \item \texttt{Conjunction}: and, with
\end{itemize}

The results of the analysis can be found in Table ~\ref{tab:lmer_results}.

\begin{table*}[ht]
\centering
\resizebox{\textwidth}{!}{%
\begin{tabular}{lcccccc}
\hline
\textbf{Predictor} & \textbf{$\hat{\beta}$} & \textbf{SE} & \textbf{df} & \textbf{\textit{t}} & \textbf{\textit{p}} & \\
\hline
\multicolumn{7}{l}{Fixed Effects} \\
\hline\hline
Intercept                                          &  0.328 & 0.055 & 130.74 &  5.984 & $< .001$ & *** \\
Gender [same]                                      &  0.092 & 0.052 & 323.11 &  1.751 & .081     & .   \\
Conjunction [with]                                 & -0.669 & 0.064 & 322.99 & -10.423 & $< .001$ & *** \\
Ont. Sim. [N + O]                    & -0.855 & 0.070 & 324.43 & -12.282 & $< .001$ & *** \\
Ont. Sim. [N + R]                  & -0.199 & 0.053 & 325.93 & -3.771 & $< .001$ & *** \\
Gender [same] $\times$ Conjunction [with]          & -0.154 & 0.074 & 323.57 & -2.083 & .038     & *   \\
Conjunction [with] $\times$ Ont. sim. [N + O]     &  0.570 & 0.099 & 327.55 &  5.739 & $< .001$ & *** \\
Conjunction [with] $\times$ Ont. sim. [N + R]   &  0.081 & 0.075 & 326.21 &  1.086 & .278     &     \\
\hline
\multicolumn{7}{l}{\textit{Random Effects}} \\
\hline
VP2 (Intercept) variance    & \multicolumn{2}{c}{0.024} & \multicolumn{2}{c}{SD = 0.155} & & \\
Residual variance             & \multicolumn{2}{c}{0.092} & \multicolumn{2}{c}{SD = 0.304} & & \\
\hline
\multicolumn{7}{l}{\textit{Model Fit}} \\
\hline
Observations                  & \multicolumn{6}{l}{350} \\
Groups (VP2)                & \multicolumn{6}{l}{27} \\
REML criterion                & \multicolumn{6}{l}{224.4} \\
\hline\hline
\multicolumn{7}{l}{\footnotesize Note: Reference categories: Gender = \textit{different}; Conjunction = \textit{and};} \\
\multicolumn{7}{l}{\footnotesize Ontological similarity = \textit{N + N}} \\
\end{tabular}
}
\caption{Linear Mixed-Effects Model Predicting Probability Difference (\textit{prob\_diff})}
\label{tab:lmer_results}
\end{table*}

\begin{table*}[t]
\centering
\resizebox{\textwidth}{!}{%
\begin{tabular}{clcc}
\hline
\textbf{Conjunction} & \textbf{Gender} & \textbf{Ontological Similarity} & \textbf{Prompt} \\
\hline
\hline
and & same & N + N & When John saw Mary and Emma, he waved at \_\_  \\
 &  & N + R & When John saw Mary and her sister, he waved at \_\_ \\
  & different & N + N & When John saw Mary and David, he waved at \_\_ \\
& & N + R & When John saw Mary and her brother, he waved at \_\_\\
& different & N + O & When John saw Mary and her bike, he waved at \_\_\\
 
\hline

with & same & N + N & When John saw Mary with Emma, he waved at \_\_  \\
 &  & N + R & When John saw Mary with her sister, he waved at \_\_ \\
  & different & N + N & When John saw Mary with David, he waved at \_\_ \\
& & N + R & When John saw Mary with her brother, he waved at \_\_\\
& different & N + O & When John saw Mary with her bike, he waved at \_\_\\

\hline

\end{tabular}
}
\caption{Summary of the dataset used in the statistical analysis. The table shows examples of the three independent variables: Gender (same, different), Ontological Similarity (N+N (name + name), N+R (name + relation), N+O (name + object)), and Conjunction (and, with)}
\label{data_t}
\end{table*}

\begin{table*}[t]
\centering
\resizebox{\textwidth}{!}{%
\begin{tabular}{lp{0.55\textwidth}p{0.35\textwidth}}
\toprule
\textbf{ID} & \textbf{Template} & \textbf{Example prompt}\\
\midrule
Original & When [S] + [$V_1$] + [$A_1$] and [$A_2$], [S] + [$V_2$] & When John saw Mary and her sister, he waved at \_\_\\
Var.~1 & Yesterday, [S] + [$V_1$] + [$A_1$] and [$A_2$] and [$V_2$] & Yesterday, John saw Mary and her sister and waved at \_\_\\
\bottomrule
\end{tabular}%
}
\caption{Sentence templates used for generating $D_{pl}$ and $D_{sg}$, with one example per template.}
\label{data}
\end{table*}

\section{Preference for Plural Pronoun is Reflected in Attention Pattern of the Pronoun Selection Head}   \label{c}

In the previous section, we show that the \textbf{Pronoun Selection} heads attend to the antecedents for the plural pronoun. To test how the activity of this head differs when the model gives a higher probability to the singular or plural pronoun, we investigate the relationship between the attention weight from the last token to $e_1$ and $e_2$ and the probability assigned to $P_{sg}$ and $P_{pl}$ in $D_{pl}$ and $D_{sg}$. We hereby refer to the last token position as \textit{i}. Let $\alpha_{i,e_{1}}$ and $\alpha_{i,e_{2}}$ be the attention weight between the last token \textit{i} and the first ($e_1$) and between \textit{i} and the second ($e_2$) antecedent of the plural construction. We define $\alpha_{diff}$ as the difference between the values of these attention weights. If $\alpha_{diff}$ is negative, $e_2$ receives more attention than $e_1$.

\vspace{0.2cm}

\begin{center}
    $\alpha_{diff} = \alpha_{i,e_{2}} - \alpha_{i,e_{1}} $
\end{center}

\vspace{0.2cm}

We also define $P_{diff}$ as the probability difference between the plural ($r_{pl}$) and singular ($r_{sg}$) pronoun.

\vspace{0.2cm}

\begin{center}
    $P_{diff} = P_{pl} - P_{sg}$
\end{center}

\vspace{0.2cm}

\begin{figure}[h]
    \centering
    \begin{subfigure}{0.48\linewidth}
        \centering
        \includegraphics[width=\linewidth]{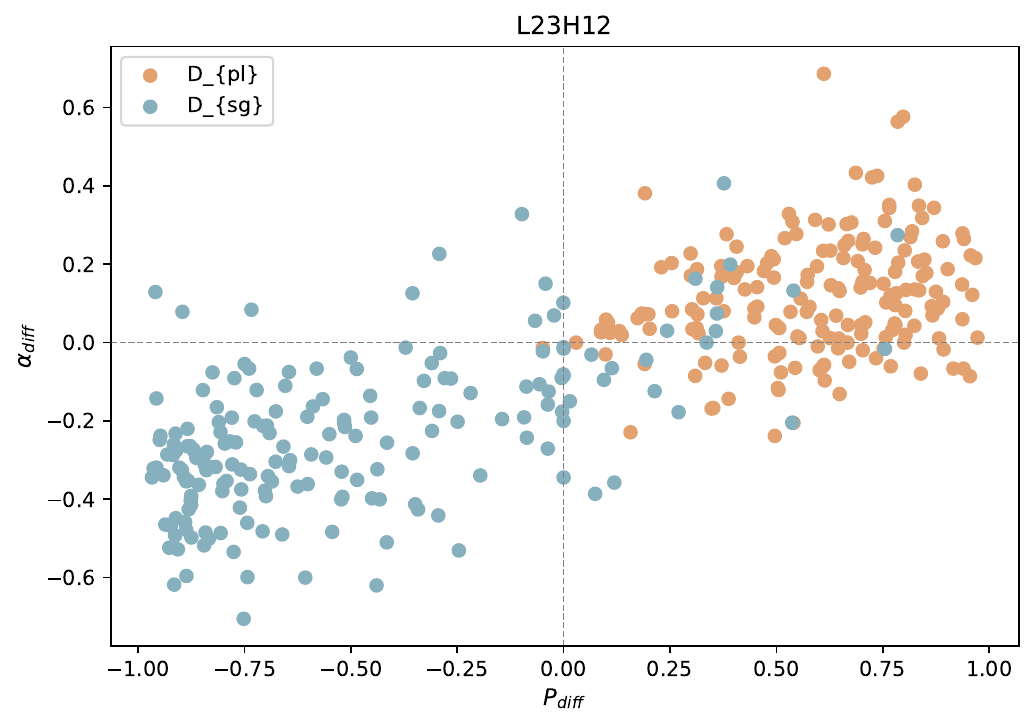}
        \caption{L23H12}
        \label{fig:first_l}
    \end{subfigure}
    \hfill
    \begin{subfigure}{0.48\linewidth}
        \centering
        \includegraphics[width=\linewidth]{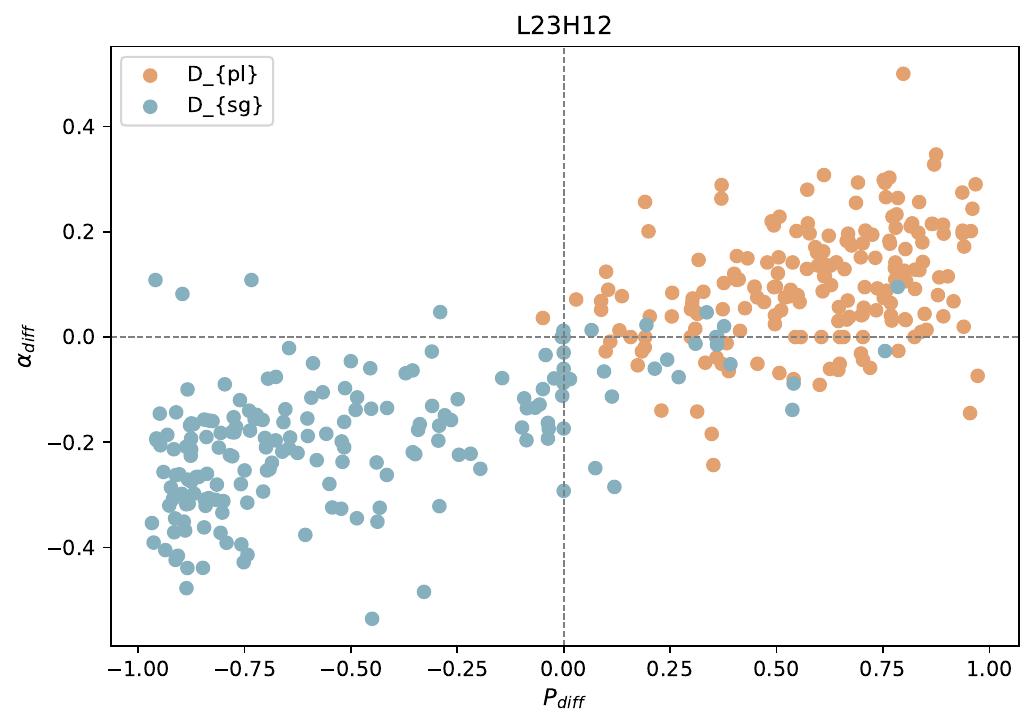}
        \caption{L23H13}
        \label{fig:second_l}
    \end{subfigure}
    \caption{Scatter plots illustrating the relationship between attention weight difference between $e_1$ and $e_2$ and difference between $P_{sg}$ and $P_{pl}$. \textcolor[HTML]{86B0BD}{Blue} dots represent data from $D_{sg}$ and \textcolor[HTML]{E2A16F}{brown} dots represent data from $D_{pl}$.}
    \label{fig:both3}
\end{figure}

We run Pearson's \textit{r} correlation on $\alpha_{diff}$ and $P_{diff}$. If the attention weight difference between $e_1$ and $e_2$ is a part of the model's decision on the probability of the singular and plural pronoun (Figure ~\ref{fig:both3}), we expect that the larger the difference between $\alpha_{i,e_{2}}$ and $\alpha_{i,e_{1}}$, the larger the difference between $P_{pl}$ and $P_{sg}$.

We found that there is a possible correlation between $\alpha_{diff}$ and $P_{diff}$ for both \texttt{L23H12} ($r$ = .78, $p$ < .001) and \texttt{L23H13} ($r$ = .77, $p$ < .001). This suggests that when the Pronoun Selection heads attend more strongly to $e_2$ relative to $e_1$, the model assigns a higher probability to the plural pronoun over the singular. This provides further evidence that the attention pattern of the Pronoun Selection head toward the two entities is a direct reflection of the model's pronoun choice.

\section{Generalization to Other Templates} \label{d}

\begin{figure*}[ht]
    \centering
    \includegraphics[width=\linewidth]{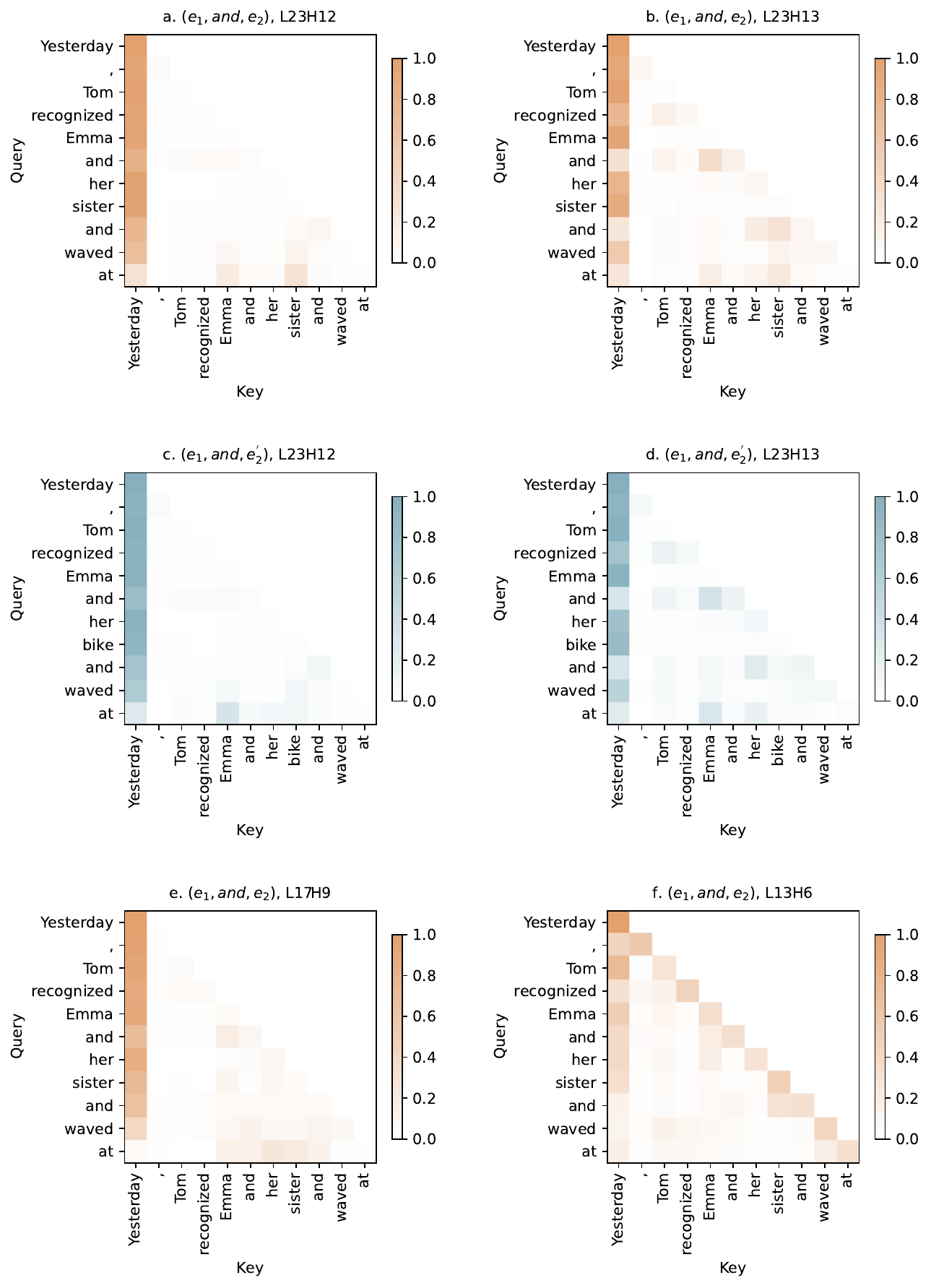}
    \caption{\textbf{Attention pattern visualizations of major heads for \texttt{Qwen3-1.7B} (Var. 1).} (a) and (b) Attention pattern of Pronoun Selection head for $D_{pl}$ (\textcolor[HTML]{E2A16F}{brown}). (c) and (d) Attention pattern of Pronoun Selection heads for $D_{sg}$ (\textcolor[HTML]{86B0BD}{blue}). (e) Attention pattern of Plurality Formation head for $D_{pl}$ (\textcolor[HTML]{E2A16F}{brown}). (f) Attention pattern of Pronoun Interpretation head for $D_{pl}$ (\textcolor[HTML]{E2A16F}{brown}).  }
    \label{fig:main3}
\end{figure*}

Table ~\ref{data} shows the original prompt template and its variant. Figure ~\ref{fig:main3} shows the attention patterns for the variant at all heads in the circuit.  

\section{Generalization to Other Models} \label{e}

\begin{figure*}[ht]
    \centering
    \includegraphics[width=\linewidth]{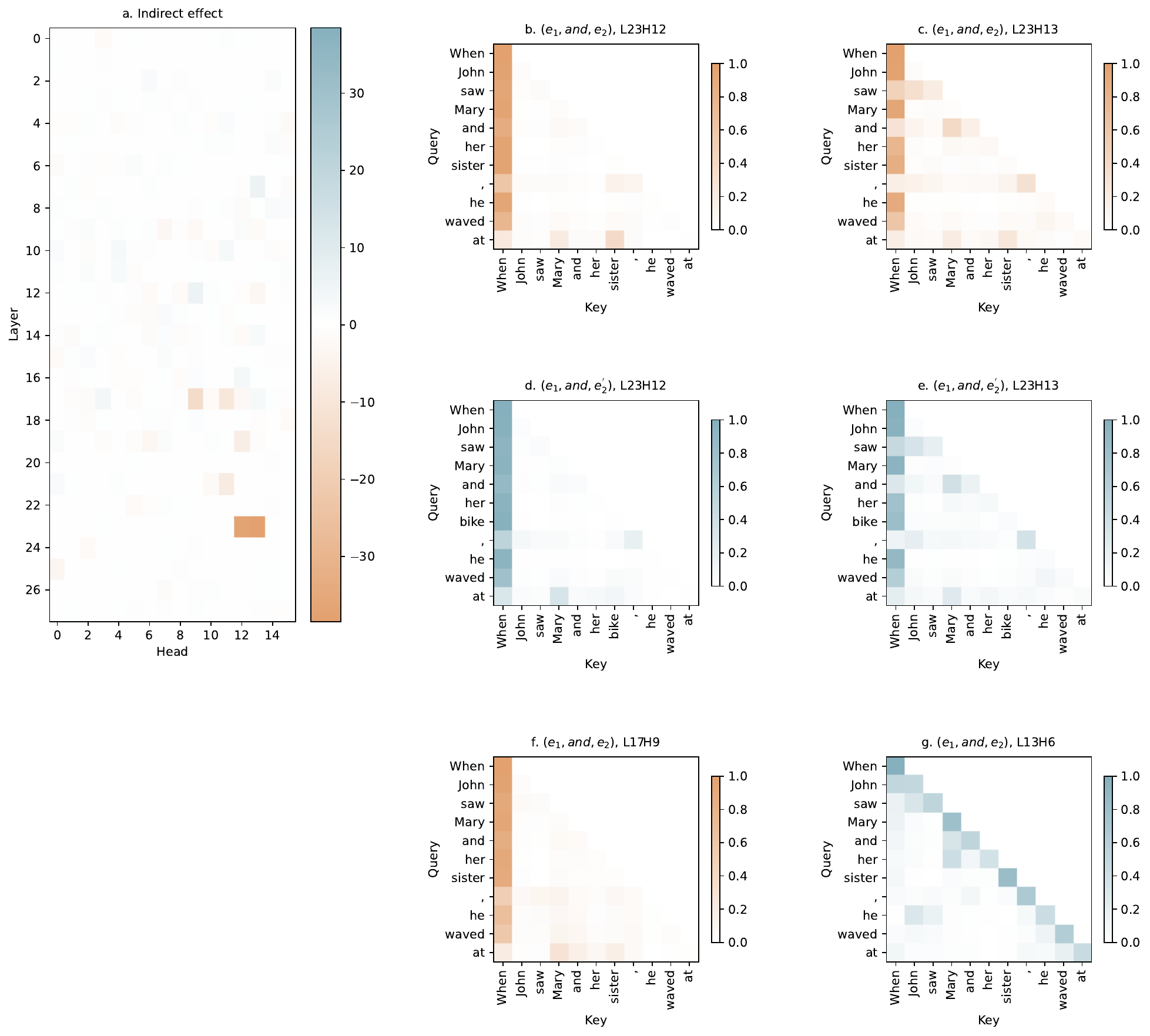}
    \caption{\textbf{Overview of intervention results and attention pattern visualizations of major heads for \texttt{Qwen3-0.6B}.} (a) Indirect effect of individual head intervention across layers. \textcolor[HTML]{E2A16F}{Brown} values denote negative change in $P_{pl}$. (b) and (c) Attention pattern of Pronoun Selection heads for $D_{pl}$ (\textcolor[HTML]{E2A16F}{brown}). (d) and (e) Attention pattern of Pronoun Selection heads for $D_{sg}$ (\textcolor[HTML]{86B0BD}{blue}). (f)  Attention pattern of Plurality Formation head for $D_{pl}$ (\textcolor[HTML]{E2A16F}{brown}). (g) Attention pattern of Pronoun Interpretation head for $D_{pl}$ (\textcolor[HTML]{E2A16F}{brown}). }
    \label{fig:0.6}
\end{figure*}

\begin{figure*}[ht]
    \centering
    \includegraphics[width=\linewidth]{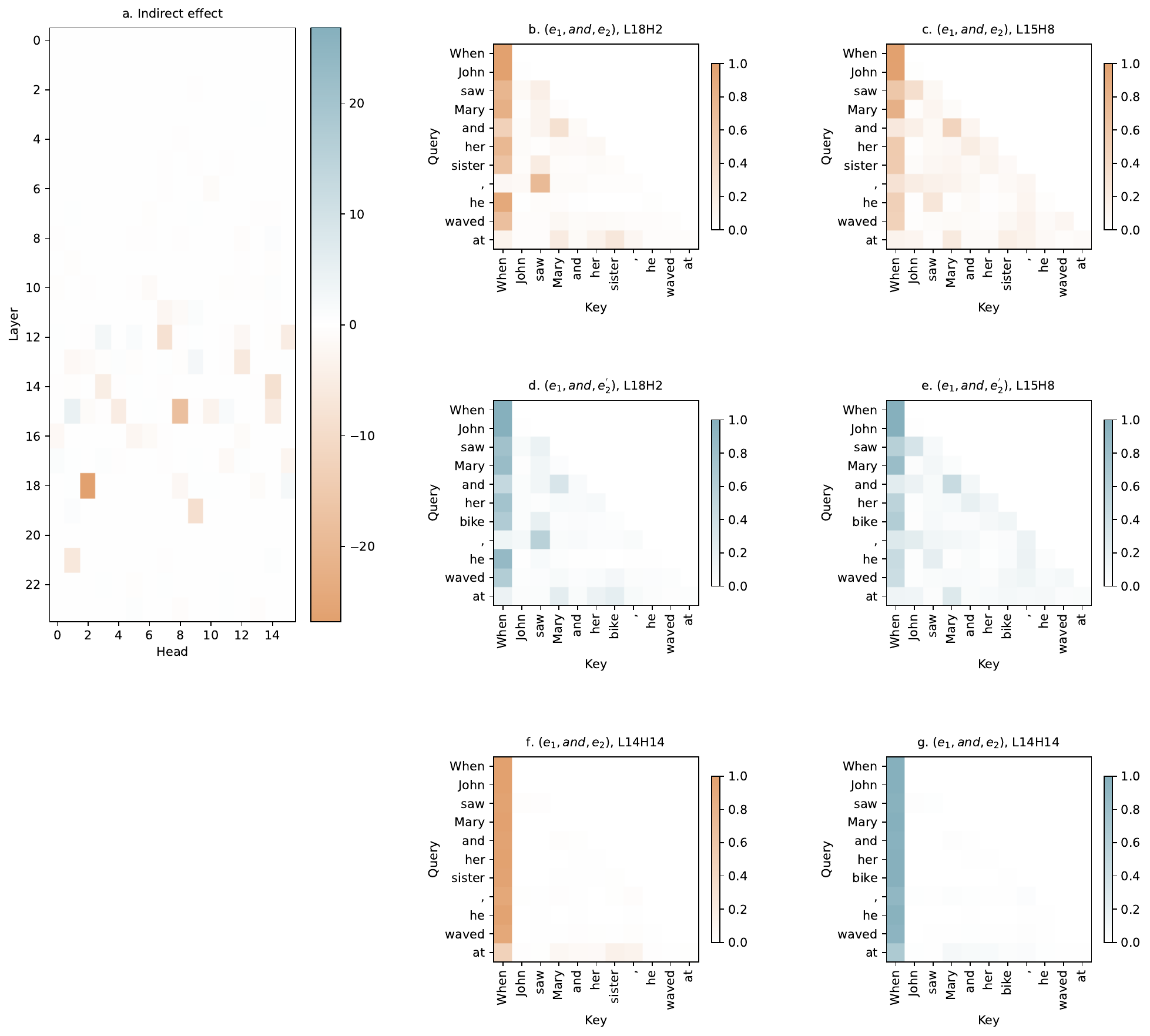}
    \caption{\textbf{Overview of intervention results and attention pattern visualizations of major heads for \texttt{GPT2-medium}.} (a) Indirect effect of individual head intervention across layers. \textcolor[HTML]{E2A16F}{Brown} values denote negative change in $P_{pl}$. (b) and (c) Attention pattern of Pronoun Selection heads for $D_{pl}$ (\textcolor[HTML]{E2A16F}{brown}). (d) and (e) Attention pattern of Pronoun Selection head for $D_{sg}$ (\textcolor[HTML]{86B0BD}{blue}). (f) and (g)  Attention pattern of Plurality Formation head for $D_{pl}$ (\textcolor[HTML]{E2A16F}{brown}) and $D_{sg}$ (\textcolor[HTML]{86B0BD}{blue}). }
    \label{fig:gpt2}
\end{figure*}

Figures ~\ref{fig:0.6} and ~\ref{fig:gpt2} show the attention patterns for the Plural Reference Circuit of \texttt{Qwen3-0.6B} and \texttt{GPT2-medium}.

\end{document}